\documentclass[11pt]{article}

\usepackage[utf8]{inputenc}
\usepackage[T1]{fontenc}
\usepackage{lmodern}
\usepackage{microtype}

\usepackage[a4paper, top=2.5cm, bottom=2.5cm, left=2.8cm, right=2.8cm]{geometry}

\usepackage{amsmath}
\usepackage{amssymb}

\usepackage{booktabs}
\usepackage{longtable}
\usepackage{array}
\usepackage{multirow}
\usepackage{calc}

\usepackage{graphicx}
\usepackage{xcolor}

\usepackage[
  colorlinks=true,
  hyperfootnotes=false,
  linkcolor=blue!70!black,
  citecolor=blue!70!black,
  urlcolor=blue!70!black,
  pdfauthor={Bo Zou and Chao Xu},
  pdftitle={Creative Quality Alignment: Expert Tacit Knowledge Transfer
            via Chain-of-Thought Fine-Tuning},
]{hyperref}
\usepackage{bookmark}

\usepackage{natbib}
\usepackage{parskip}
\usepackage{setspace}
\title{\textbf{Creative Quality Alignment:\\
Expert Tacit Knowledge Transfer via Chain-of-Thought Fine-Tuning}}
\author{Bo Zou \and Chao Xu}
\date{2026}

\newcounter{none}
\providecommand{\tightlist}{\setlength{\itemsep}{0pt}\setlength{\parskip}{0pt}}

\begin{document}
\maketitle

\begin{abstract}

This paper provides an empirical implementation of the creative quality
metric proposed in \emph{Calibrated Surprise}
\citep{calibratedsurprise2026}. That paper defines literary quality in
terms of mutual information \(I(X;Y) = H(X) - H(X|Y)\), and shows that
the degree to which \(H(X|Y)\) is reduced under the full-dimensional
constraint
\(Y = Y_{\text{intent}} \cap Y_{\text{audience}} \cap Y_{\text{reality}}\)
reflects how sensitive a model is to that constraint. The question this
paper addresses is: does this mathematical claim hold at the engineering
level?

To make the answer as general as possible, we deliberately choose the
strictest engineering conditions: low data cost and a small base model.
If \(H(X|Y)\) can be significantly reduced here, larger-scale
implementations will follow more easily. Specifically: training data
comes from approximately 100 expert chain-of-thought (CoT) annotations
produced by the BC Protocol \citep{bc2026}. The base model is
Qwen2.5-7B-Base. Calibration is measured on an independent benchmark
\citep{bench2026} using two objective, reproducible quantities: the
change in conditional mutual information \(\Delta(\Delta I)\) as the
primary metric, and per-token logprob differences \(\Delta\log P\) as an
auxiliary view. Neither relies on human scoring.

Main findings: at epoch 5, the model's sensitivity to literary
constraints---measured by \(\Delta(\Delta I)\), the change in
conditional mutual information before and after fine-tuning---shows a
statistically significant improvement (literary group, \(n=20\) paired,
Wilcoxon one-sided \(p = 0.041\), Cohen's \(d = 0.50\)). On the same
checkpoint, the factual control group shows no significant effect
(\(p = 0.689\), \(d = -0.12\)). This contrast is direct evidence that
\(P(x|y)\) calibration is \textbf{domain-selective}. We also identify a
data bias: most publicly available alignment datasets are skewed toward
craft-related knowledge, while audience modeling and reality-logic
coverage are systematically weak---two of the three constraint types are
not explicitly taught.

We use the term Creative Quality Alignment (CQA) to describe this class
of engineering methods---those that extend from \emph{Calibrated
Surprise} toward aligning LLM creative judgment. CQA does not introduce
a new alignment mechanism. The foundational contributions come from
\emph{Calibrated Surprise} (mathematical metric) and the BC Protocol
paper (data production methodology). CQA identifies the first
engineering path that brings both together. We also offer a supporting
theoretical observation: in an LLM with a single conditional
distribution architecture, appreciation and generation are two forward
passes of the same \(P_\theta\). Calibrating the appreciation side
automatically transfers to the generation side via architectural
duality. This is the structural reason why $\sim$100 CoT
examples are sufficient---not a purely empirical observation like
LIMA~\citep{zhou2023lima}.

\end{abstract}

\section{Introduction}\label{introduction}

\subsection{The Problem: Structural Limitations of Rubric-Based Creative Evaluation}\label{the-problem-structural-limitations-of-rubric-based-creative-evaluation}

Alignment pipelines deployed at scale give an implicit answer to the
question of how to measure creative judgment: the rubric paradigm. It
decomposes creative quality into discrete sub-dimensions, scores each
independently, and aggregates via weighted sum. Rubrics-as-Rewards is
one representative industrial implementation. This approach provides a
quantifiable annotation interface, but it encodes a strong mathematical
assumption: the evaluation dimensions are conditionally independent.
Under this assumption, the joint probability
\(P(x|y_1, y_2, \ldots, y_n)\) is approximated by the product of
marginals \(\prod_i P(x|y_i)\).

In real creative work, the dimensions are tightly coupled. Character
psychology shapes causal logic. Causal logic constrains tone. Tone in
turn defines thematic expression. Any linear weighting
\(S = \sum_i w_i \cdot s_i\) systematically omits cross-dimension
interaction terms, regardless of how the weights are set. This parallels
a basic fact in statistics: knowing all marginal distributions is not
enough to recover the joint distribution.

Recent empirical work confirms this independently. Rubric evaluation
fixes its dimension set at design time, while high-quality creative work
can demonstrate value on dimensions the evaluator never anticipated
\citep{messick1994}. LLM evaluators under rubric decomposition also
confuse quality dimensions and cannot reliably distinguish independent
sub-items \citep{hu2024llm,howcroft2020,kaufman2008}.

The core problem is clear. In subjective domains like creative judgment,
quality does not come from sub-item scores after decomposition. It comes
from holistic judgment under full-dimensional constraints. Our own team
went through a similar path early in this project---designing a
multi-level tree-structured annotation scheme (§3.5) that was an extreme
version of the rubric paradigm---and it failed for the same reasons.

\subsection{Research Question and Scope}\label{research-question-and-scope}

\emph{Calibrated Surprise} \citep{calibratedsurprise2026} provides a
formal definition of literary quality: \(I(X;Y) = H(X) - H(X|Y)\), where
quality improves as \(H(X|Y)\) is reduced under the full-dimensional
constraint. This result is mathematically sound. If its premises are
accepted, the conclusion follows necessarily, independent of empirical
data. (\emph{Calibrated Surprise} §7 includes 20 sanity-check
calculations, but these are post-hoc confirmations, not the logical
basis of the argument.)

Whether a mathematically valid claim holds at the engineering level is a
separate question. That is what this paper addresses. Can a model's
\(H(X|Y)\) actually be reduced? Does training data need to cover all
three constraint types---\(Y_{\text{intent}}\), \(Y_{\text{audience}}\),
\(Y_{\text{reality}}\)---at the corpus level? And can this reduction be
verified with objective, reproducible metrics that do not depend on
human judgment?

To make the answer as general as possible, we deliberately use the
strictest conditions: $\sim$100 samples, LoRA-based supervised
fine-tuning, and a small base model. If the reduction is significant
under these conditions, larger-scale implementations will be easier to
achieve.

We call this class of engineering methods Creative Quality Alignment
(CQA). The term does not refer to a new alignment mechanism. It labels
the class of engineering methods that extend from \emph{Calibrated
Surprise} toward aligning LLM creative judgment. This paper describes
one concrete path; future implementations from the same theoretical base
also belong to CQA. The foundational contributions come from
\emph{Calibrated Surprise} (mathematical metric) and the BC Protocol
paper (data production methodology).

One terminological note. We do not use phrases like ``improving the
model's aesthetic judgment'' or ``improving creative ability.'' These
are vague at the mechanism level and cannot be mapped to independently
verifiable metrics. Our precise formulation throughout is: the
sensitivity of the model's internal conditional distribution \(P(x|y)\)
to the full-dimensional constraint \(Y\) is increased, which is
equivalent to reducing the conditional entropy \(H(X|Y)\). This can be
observed directly from logprobs, without human scoring.

This choice has an additional benefit. In an LLM, \(H(X|Y)\)
characterizes both the appreciation and generation sides at once.
Generation is sampling from \(P_\theta(x|y)\). Appreciation is reading
the logprob from \(P_\theta(x|y)\). These are not two separate
capabilities---they are two forward passes of the same object. By the
symmetry \(I(X;Y) = H(X) - H(X|Y) = H(Y) - H(Y|X)\), optimizing
\(H(X|Y)\) calibrates both sides simultaneously. The full argument is in
§6.5, but the reason for choosing this proxy metric is already complete
here.

\subsection{Three Types of Tacit Knowledge}\label{three-types-of-tacit-knowledge}

Once the training objective is clear, the question becomes: what types
of knowledge need to be injected? Our observation is that high-quality
creative judgment CoT must carry three types of tacit knowledge. Each
type corresponds to one of the three constraint sources in the
\emph{Calibrated Surprise} framework.

\textbf{Craft and aesthetic knowledge}---corresponding to
\(Y_{\text{intent}}\). This covers judgments from the creator's
perspective: ``why this technique is effective,'' ``why this dialogue is
better than an alternative.''

\textbf{Audience psychology modeling}---corresponding to
\(Y_{\text{audience}}\). This covers judgments from the reader's
perspective: ``how will the reader's expectations shift at this point,''
``is the timing of this information disclosure appropriate.''

\textbf{Reality logic mapping}---corresponding to
\(Y_{\text{reality}}\). This covers judgments about real-world
constraints: ``how would a domestic abuse survivor actually react in a
public setting,'' ``how would someone in this social situation respond
to this circumstance.''

These three types map directly to the full-dimensional constraint in
\emph{Calibrated Surprise}:

\[Y = Y_{\text{intent}} \cap Y_{\text{audience}} \cap Y_{\text{reality}}\]

``Carried simultaneously'' needs a precise definition. It does not mean
every CoT sample must explicitly address all three types. Creative
decisions vary---some mainly involve psychological realism, others are
primarily governed by causal logic. ``Simultaneously'' means the CoT
corpus as a whole must contain all three types. It is not a per-sample
coverage requirement.

This is not a trivial observation. It is a diagnostic finding about
current publicly available alignment data. In the open alignment
datasets we have examined, annotations are overwhelmingly focused on
craft knowledge---fluency, logical consistency---with almost no explicit
teaching of audience psychology, and even less of reality logic mapping.
Two of the three tacit knowledge types are systematically
underrepresented.

\subsection{Academic Precedents for Three-Dimensional Convergence}\label{academic-precedents-for-three-dimensional-convergence}

The three-way convergence is not just an intuition from creative
practice. It has independent roots in literary theory, linguistics, and
semiotics. But in the engineering context of LLM alignment and creative
evaluation, it has not previously been unified into an operational
evaluation framework.

\citet{abrams1953} in \emph{The Mirror and the Lamp} proposed a
four-element model of literary criticism: universe (objective reality),
audience, artist (author), and work. The first three---excluding the
work itself---map precisely to the three-way convergence. But Abrams's
framework is descriptive taxonomy, not a normative evaluation tool.

\citet{buhler1934}'s Organon Model identifies three irreducible
functions of the linguistic sign: expression (\emph{Ausdruck}, pointing
to author intent), appeal (\emph{Appell}, pointing to audience
response), and representation (\emph{Darstellung}, pointing to objective
reality). Bühler argues explicitly that the three functions are
inseparable---removing any one causes communicative breakdown.

Eco's three-intent theory \citep{eco1990}---\emph{intentio auctoris},
\emph{intentio lectoris}, \emph{intentio operis}---overlaps structurally
with the three-way convergence. But Eco's focus is hermeneutics (how to
interpret a work correctly), not evaluation (what structure produces
quality).

The most direct correspondence comes from cognitive writing studies.
\citet{grabe1996} in \emph{Theory and Practice of Writing} argue
explicitly that a successful text must be simultaneously constrained by
three forces: writer intent, audience expectations, and the shared
phenomenological world.

Our contribution is not in discovering these three dimensions. Prior
work already touches on the three-way structure from descriptive,
linguistic, hermeneutic, and writing perspectives. Our contribution is
unifying this scattered consensus into an operational evaluation
framework, anchored to the formal definition
\(Y = Y_{\text{intent}} \cap Y_{\text{audience}} \cap Y_{\text{reality}}\)
in \emph{Calibrated Surprise}.

\subsection{Contributions}\label{contributions}

This paper's contributions are at the empirical implementation level:

\begin{enumerate}
\def\labelenumi{\arabic{enumi}.}
\tightlist
\item
  We provide empirical support, in a low-cost post-training setting, for
  the claim that ``creative quality improves as \(H(X|Y)\) is reduced,''
  via systematic fine-tuning experiments.
\item
  Drawing on the academic precedents in §1.4, the BC Protocol probing
  design in §3.1, and the domain contrast results in §5.1, we argue that
  all three constraint types must be present in the CoT corpus as a
  whole. We use this position to diagnose the systematic
  underrepresentation of two constraint types in current open alignment
  datasets.
\item
  We provide a validation method using logprob differences
  (\(\Delta\log P\)) on an independent benchmark as an objective
  metric---no human judgment, no model self-scoring.
\item
  We unify the three-way structure from literary theory and writing
  studies \citep{abrams1953,buhler1934,eco1990,grabe1996} under the
  evaluation framework of \emph{Calibrated Surprise}.
\item
  Using our team's early multi-level tree-structured annotation scheme
  as a failed case, we present an epistemological lesson about
  decomposing holistic creative judgment into independent sub-items
  (§3.5).
\item
  We provide a supporting architectural argument (§6.5): in an LLM's
  single conditional distribution, appreciation and generation share the
  same \(P_\theta\), so calibrating the appreciation side automatically
  carries over to the generation side. This gives a structural reason
  why $\sim$100 CoT examples are sufficient, and makes the dual
  effectiveness of CQA training a mathematical prediction, not an
  empirical coincidence.
\end{enumerate}

\section{Background and Related Work}\label{background-and-related-work}

\subsection{\texorpdfstring{Key Elements of the \emph{Calibrated Surprise} Framework}{Key Elements of the Calibrated Surprise Framework}}\label{key-elements-of-the-calibrated-surprise-framework}

\emph{Calibrated Surprise} \citep{calibratedsurprise2026} uses Shannon
information theory to provide the mathematical definitions and notation
that this paper's experiments depend on. We summarize only the elements
directly relevant here.

The mutual information formula \(I(X;Y) = H(X) - H(X|Y)\) describes, in
the context of creative quality, how much a creative choice \(X\) is
driven by the full-dimensional constraint \(Y\). Higher \(I(X;Y)\) means
the choice is closer to being uniquely determined by the constraint,
rather than spread across several equally likely alternatives.

The roles of this paper, \emph{Calibrated Surprise}, and the companion
Benchmark paper can be separated as follows:

\begin{itemize}
\tightlist
\item
  \textbf{\(I(X;Y)\) measures works.} As a Shannon information metric,
  it characterizes how much a specific creative choice is forced by the
  constraint. Its target is the work. This is the core tool in
  \emph{Calibrated Surprise}.
\item
  \textbf{CQA calibrates the model's \(H(X|Y)\).} The target is the
  model's internal conditional distribution \(P(x|y)\). The goal is to
  make it more sensitive to \(Y\). This is the core task of this paper.
\item
  \textbf{The Benchmark tests whether calibration succeeded.} It uses
  known reference answers to diagnose the model's \(P(x|y)\) from the
  outside. This is the core tool in the companion Benchmark paper
  \citep{bench2026}.
\end{itemize}

The three layers are not interchangeable: they operate on different
objects.

This paper's experimental logic depends on a basic property from Shannon
information theory: \textbf{conditioning reduces entropy}.

\[H(X|Y, Z) \leq H(X|Y)\]

That is, adding any effective information \(Z\) on top of \(Y\)---here,
the conditional information carried by one new expert CoT---can only
decrease or maintain the conditional entropy, never increase it. This
property provides a mathematical guarantee: every expert CoT that
carries genuine information will reduce the model's \(H(X|Y)\) by at
least zero. There is no scenario where adding more relevant information
makes the conditional distribution flatter. The derivation is in
\emph{Calibrated Surprise} §2.2.1.

\emph{A historical note:} The two lines of work---this paper and
\emph{Calibrated Surprise}---developed in parallel and influenced each
other. Our team initially worked along a rubric-based path, which
failed. Switching to the BC Protocol and natural-language CoT led to the
mathematical framework that became \emph{Calibrated Surprise}.

\subsection{Systematic Limitations of Existing Alignment Data}\label{systematic-limitations-of-existing-alignment-data}

We briefly survey the main alignment data production methods in current
use. Each has a structural weakness at a different stage.

\textbf{Crowdsourced annotation.} Annotators are typically not domain
experts. The output is mostly shallow preference judgments with no
accompanying reasoning. Intermediate inference steps are almost never
recorded.

\textbf{Independent expert writing.} This suffers from the ``expert
blind spot'' \citep{polanyi1966,nathan2003}. Experts tend to skip
intermediate reasoning steps they consider obvious. The result is closer
to a polished conclusion than a full chain of inference.

\textbf{RLHF preference ranking.} The signal produced is a preference
pair (A is better than B), not a reasoning chain explaining why. This
format can transmit preference direction, but it is hard to transmit the
reasoning process itself.

\textbf{Synthetic data} (e.g., Self-Instruct
\citep{wang2023selfinstruct}). This is essentially the model distilling
its own existing capabilities. It cannot inject expert knowledge that
lies outside the model's current distribution.

This list is not exhaustive. DPO/IPO optimization paths, Constitutional
AI, and Rubrics-as-Rewards (already discussed in §1.1) are also
mainstream approaches, but they are less directly relevant to this
paper's methodological focus.

A common blind spot runs through all of these: they focus on the
quantity, format, and diversity of data, but rarely on the
\emph{production process}---specifically, the cognitive conditions under
which expert knowledge is elicited. The BC Protocol paper \citep{bc2026}
addresses exactly this gap. The fine-tuning data used in this paper is
natural-language CoT produced by BC dialogues.

\subsection{Additional Limitations of Rubric-Based Evaluation}\label{additional-limitations-of-rubric-based-evaluation}

§1.1 identified the core mathematical flaw of rubric evaluation: the
conditional independence assumption. Here we add two related limitations
from different angles.

\textbf{Construct underrepresentation.} Rubric evaluation fixes its
dimension set at design time. But high-quality creative work can
demonstrate value on dimensions the evaluator never anticipated.
Aggregating scores through weighted averaging cannot recover the
holistic structure that emerges from dimension interactions. Messick
(1994)'s concept of construct underrepresentation captures this
precisely: any rubric with a finite number of dimensions will, in
principle, fail to fully represent the construct it aims to measure. A
holistic aesthetic judgment---``this work is appropriate as a
whole''---cannot be reconstructed by decomposing and re-aggregating
sub-items.

\textbf{Measuring expression, not state.} When an LLM is asked to score
under a rubric, what is being measured is whether it can produce a
well-formed verbal justification---not whether its internal distribution
is actually sensitive to the relevant constraints. A model can fluently
explain ``why this psychological passage is profound'' while its
conditional distribution \(P(x|y)\) remains poorly calibrated on the
dimension of psychological realism. This is a gap between the expression
layer and the distribution layer.

\subsection{Epistemological Parallel to the Jelinek Paradigm Shift}\label{epistemological-parallel-to-the-jelinek-paradigm-shift}

The multi-level tree-structured annotation scheme our team designed
early in this project (§3.5) was, in its logic, an extreme version of
the rubric paradigm. Its failure has the same methodological structure
as the failure of hand-crafted linguistic rules in NLP history.

The IBM speech recognition team led by Jelinek completed a paradigm
shift from hand-crafted linguistic rules to statistical modeling during
the 1970s--1980s. The key move was not simply abandoning expert
knowledge. It was abandoning the practice of explicitly encoding expert
understanding of linguistic structure as programmatic rules---and
instead letting the model learn those structures implicitly from data.

Our methodological shift shares the same epistemological core: abandon
the explicit encoding of holistic aesthetic judgment into independent
sub-items (the rubric approach), and instead use natural-language CoT to
directly carry the reasoning process that experts use when making
holistic judgments (the statistical approach).

\section{Method: The CQA Pipeline}\label{method-the-cqa-pipeline}

This section follows the engineering sequence of data → resources →
training. §3.1 describes how the SNAKE-100 training data was produced
(BC Protocol dialogues). §3.2 and §3.3 cover methodological
considerations for CoT quality and decision-point selection. §3.4 gives
the LoRA fine-tuning implementation. §3.5 provides the design rationale
for using natural-language CoT rather than structured rubrics---using
our team's early failed approach as the negative case, in parallel with
the Jelinek analogy in §2.4. Readers focused on the engineering
implementation can go directly from §3.4 to Section 4; §3.5 is extended
reading for those interested in methodological motivation.

\subsection{Data Source: BC Protocol Dialogues}\label{data-source-bc-protocol-dialogues}

The training data for CQA comes from the BC Protocol \citep{bc2026}---a
structured two-expert dialogue method designed for high-density CoT
production. The core setup:

\begin{itemize}
\tightlist
\item
  \textbf{Role configuration.} B (knowledge engineer, high fluid
  intelligence) and C (domain expert, high crystallized intelligence)
  conduct paired voice dialogues.
\item
  \textbf{B's elicitation function.} B uses epistemic vigilance to probe
  the implicit reasoning steps in C's judgments, making them explicit
  and recordable as natural-language inference chains.
\item
  \textbf{Typical output.} Approximately 10--20 fine-tuning-ready CoT
  samples per hour of dialogue.
\end{itemize}

The data production is shared with the BC Protocol paper: the same BC
dialogues are the joint output of both papers, analyzed from two
different angles---the BC Protocol paper focuses on the data production
methodology itself; this paper focuses on the downstream fine-tuning
effects.

We name the $\sim$100 expert CoT dataset used for CQA
fine-tuning \textbf{SNAKE-100}. The name comes from the SNAKE mechanism
used in BC dialogues to elicit tacit knowledge (see BC Protocol paper
§3.2.4). This name provides a consistent label for citation and dataset
release.

One potential misunderstanding needs to be addressed: the
three-dimensional convergence
(\(Y_{\text{intent}} \cap Y_{\text{audience}} \cap Y_{\text{reality}}\))
is not an annotation protocol in BC dialogues. It is not a requirement
that each sample must explicitly cover all three dimensions. It is the
radar that B (the questioner) carries internally---when C's judgment
explicitly addresses one dimension while implicitly skipping another, B
asks a follow-up to check whether the skipped dimension also holds. In
other words, the three dimensions live in the questioner's epistemic
vigilance, not in any surface-level dialogue protocol.

The reason the three dimensions cannot be turned into a protocol is
mathematical. When C makes a high-quality creative judgment, all three
dimensions have already converged into a single intersection judgment in
C's holistic cognition. Requiring C to report each dimension separately
means linearizing this already-converged judgment into three parallel
marginal projections---exactly the decomposition that §1.1 rejects
mathematically (and whose extreme form the multi-level tree annotation
scheme demonstrated empirically as a failure; see §3.5). Reducing to 3
dimensions does not fix this mathematical flaw; it only hides it better.

The signal for when B can stop probing a sample is also controlled by B,
not C. If C decided when to stop, the method would degrade back to the
single-expert writing blind spot described in §2.2---experts stop when
something seems obvious to them. B's stopping criterion is different:
stop when genuine uncertainty is exhausted and further probing would
only produce confabulation, not new judgment.

Two observable consequences follow. First, the actual length of a single
CoT is determined by the true complexity of the judgment---it may be
significantly longer than typical alignment data samples. This length is
handled downstream by the engineering infrastructure (multi-GPU, long
context); it should not be compressed at the dialogue stage. Second, the
precision of B's stopping point is a hidden quality bottleneck: the
actual information density of 100 CoT samples partly depends on how well
B judges when to stop on each one. That judgment cannot be reduced to
any explicit protocol. This point reappears in §6.6.

\subsection{Data Quality}\label{data-quality}

This section states the methodological stance on quality analysis for
the CoT samples produced by BC dialogues.

\textbf{Overall principle.} We follow the Selection-over-Prescription
stance consistent with the BC Protocol: quality assurance is
concentrated in personnel selection and the dialogue process, not in
filtering data using explicit coverage requirements after the fact.

In practice, all CoT samples produced naturally in BC dialogues are
kept. Samples are not excluded for ``only touching one dimension'' or
``failing to cover a given dimension.'' The reason is that filtering by
a preset dimension set is exactly the same rubric logic applied to data
selection---after that logic has already been rejected at the scoring
level in §1.1, using it at the filtering level would be inconsistent.

We instead use post-hoc descriptive statistics to characterize the
actual distribution of data: we record, for each CoT, which of the three
tacit knowledge types it touches (``all three / two / one''), as a
descriptive count. The full distribution for SNAKE-100 will be released
with the dataset in arXiv v2. This count is presented as an objective
description of the data's shape, not as a target the data should conform
to.

A natural follow-up would be a post-hoc slice ablation: drop samples
touching only one or two dimensions, retrain, and check whether
\(\Delta(\Delta I)\) drops. We do not run that ablation for the reasons
given in §4.3. The descriptive count above is therefore reported as data
documentation, not as a filtering criterion that the corpus needs to
satisfy.

\subsection{The Lossy Compression Perspective}\label{the-lossy-compression-perspective}

Any process of making tacit expert knowledge explicit is, in principle,
lossy. The expert's internal judgment system carries far more
information than any CoT transcript can hold. The methodological
difference is not whether there is loss, but whether the location of the
loss is actively chosen: passive loss means not knowing what was
dropped; active loss means using methodological judgment to choose which
information to sacrifice.

We actively exclude three types of knowledge with lower information
gain:

\textbf{Pure sensory intuition.} Judgments like ``this sentence doesn't
read right.'' These cannot in principle be made explicit as reusable
reasoning chains, so they offer almost no learnable signal for an LLM.

\textbf{Highly personal preferences.} Judgments like ``the author
personally dislikes second-person narration.'' These reflect individual
aesthetic taste and do not generalize across samples.

\textbf{Pure procedural knowledge.} Judgments like ``write the ending
before the opening.'' These describe how a creative workflow is
organized, not the reasoning behind a creative choice. They do not
directly contribute to calibrating \(P(x|y)\).

Actively declaring which types are excluded---and why---has
methodological value in itself. The pattern is shared with other
domains: JPEG actively discards high-frequency components that human
vision is less sensitive to; the bias-variance tradeoff in statistics
actively introduces bias to reduce estimation variance. All three cases
share the same form: given unavoidable loss, choosing where the loss
occurs.

Among what is kept, which type of CoT information most benefits LLM
training? Our observation is that decision-turn reasoning---where the
reasoning has the form ``initially I judged X, but given Y, the final
choice is Z''---has significantly higher learning efficiency than purely
descriptive factual statements. The reason is that decision-turn
reasoning directly encodes conditional dependencies in its surface
structure---which is exactly the signal CQA depends on to calibrate
\(P(x|y)\).

\subsubsection{A Logprob-Based Training Efficiency Strategy}\label{a-logprob-based-training-efficiency-strategy}

The three exclusion criteria in §3.3 come from methodological
experience. This section proposes a data-driven auxiliary strategy to
further allocate training attention among the retained CoT samples.

\textbf{Core criterion.} The CoT information that contributes most to
reducing \(H(X|Y)\) is that which encodes conditional dependencies that
the pretrained model's existing knowledge cannot already derive. The
marginal value of an expert CoT depends on the incremental information
it carries relative to the pretrained distribution.

\textbf{Operational steps:}

\begin{enumerate}
\def\labelenumi{\arabic{enumi}.}
\tightlist
\item
  For each narrative decision point, compute the logprobs from the
  pretrained base model (before CQA fine-tuning) to read the model's
  current conditional distribution.
\item
  If the model already assigns high probability to the correct choice,
  the judgment at that decision point has been substantially
  internalized during pretraining, and the marginal benefit of teaching
  it with expert CoT is low.
\item
  If the model's distribution is near-flat across candidates, or assigns
  high probability to an incorrect choice, that decision point is where
  expert CoT should be concentrated.
\end{enumerate}

\textbf{Formal statement.} The marginal value of a CoT sample, relative
to the general pool of expert CoT, is proportional to:

\[D_{KL}\bigl(P_{\text{expert}}(x|y)\,\|\,P_{\text{pretrained}}(x|y)\bigr)\]

That is, the KL divergence between the expert conditional distribution
and the pretrained model's conditional distribution. The larger this
divergence, the more the model deviates from the expert at that decision
point, and the higher the calibration value of the corresponding CoT.

With this strategy, the choice of ``where to accept information loss''
can be driven directly by data rather than by experience-based
assumptions about which knowledge is less valuable.

\subsection{Fine-Tuning Setup}\label{fine-tuning-setup}

\textbf{Base model.} Qwen2.5-7B-Base. This model is consistent with the
same-generation base used in the computational verification experiments
of \emph{Calibrated Surprise} §7 (the actual version was updated from
the Qwen1.5 series to the Qwen2.5 series to ensure reproducibility and
community accessibility during the 2025--2026 submission period).

\textbf{Training method.} Supervised fine-tuning using LoRA
\citep{hu2022lora} on SNAKE-100. LoRA configuration: rank=32, alpha=32,
target modules covering \{q, k, v, o, gate, up, down\}\_proj (7 modules
total); rsLoRA \citep{kalajdzievski2023} is used to correct the scaling
bias at high rank. Hyperparameters: learning rate \(2\times10^{-4}\),
cosine schedule with warmup ratio 0.1, 12 epochs total, effective batch
size 2 (single-card batch size 1 × gradient accumulation 2).
Regularization: lora\_dropout=0.05, NEFTune \citep{jain2023neftune}
noise amplitude \(\alpha=5\). Mixed precision: bf16. Hardware: 1 ×
NVIDIA A800-80 GB, training time $\sim$5 min 35 sec.~Training
framework: LLaMA-Factory \citep{zheng2024llamafactory}.

\textbf{Scale effect analysis design.} Separate training subsets of 10,
30, 60, and 100 CoT samples are used to fine-tune independently,
measuring how the effect varies with training data size. This experiment
is listed in §4.4 and will be included as arXiv v2 content.

\subsection{Design Rationale: From Multi-Level Tree Annotation to Natural-Language CoT}\label{design-rationale-from-multi-level-tree-annotation-to-natural-language-cot}

We acknowledge that the rubric paradigm played a foundational role in
the early engineering of scalable alignment annotation. Scalable
annotation requires a standardized process that distributed workers can
execute---decomposing quality into discrete, quantifiable sub-items is a
reasonable engineering choice under that constraint.

Our team made a similar choice early in this project, under the same
inertia. The scheme we designed was a multi-level tree-structured
annotation protocol: 6 top-level dimensions, each decomposed into 2--3
second-level sub-dimensions, some of which decomposed further into 2--3
third-level sub-dimensions. An annotation workflow was paired with this
structure, requiring annotators to score each decision point along the
hierarchy. This scheme was, in its logic, an extreme version of the
rubric paradigm---the tree structure meant far more dimensions than a
typical rubric, and the workflow provided strict distributed-execution
protocols.

The scheme failed in practice.

The failure had the same root cause as the mathematical flaw described
in §1.1: it assumed that holistic aesthetic judgment can be losslessly
decomposed into independent per-sub-item scores, but the decomposition
process itself already destroys the integrity of the judgment. When
scoring across many dimensions one by one, annotators found that a
holistic judgment (``this paragraph does not work as a whole'') could
not be cleanly attributed to any single sub-dimension; forcing
attribution produced outputs that were closer to distortion than
preservation.

This lesson needs to be stated at the methodological level. Otherwise it
is easy to misread as ``the number of dimensions was too large.'' The
actual failure mechanism is not the number of dimensions---it is the
\textbf{protocol structure} of ``preset dimension set + per-sample
coverage requirement.'' Whether the dimension count is 30, 5, or even 3
(exactly matching the three-dimensional convergence framework of this
paper), if the protocol requires each sample to be reported dimension by
dimension, it will linearize the expert's already-converged intersection
judgment into parallel marginal projections. The mathematical argument
in §1.1---that the joint probability \(P(x|y_1, \ldots, y_n)\) \(\neq\)
the product of marginals \(\prod_i P(x|y_i)\)---applies equally to all
such protocols. The multi-level tree scheme was just the most extreme
version of this error; the dimension count grew exponentially with
depth. Reducing to 3 dimensions does not fix the mathematical flaw; it
only makes it less visible. This is the explicit basis for the
distinction in §3.1 between ``interrogation prior'' (which preserves the
integrity of the intersection judgment) and ``coverage protocol'' (which
is mathematically lossy regardless of dimension count).

The methodological shift came when we switched to BC Protocol
natural-language dialogues as the elicitation medium. Under this
approach, experts are no longer asked to decompose their judgments by
preset dimensions. Instead, they present their full reasoning in natural
language---the typical form being: ``My judgment here is that this
doesn't work, because\ldots; but I also have to consider\ldots; so the
right choice is\ldots.'' This reasoning chain carries the
cross-dimensional constraints simultaneously in its surface structure,
and also preserves the couplings between dimensions.

This shift shares the same epistemological core as the Jelinek team's
transition from hand-crafted rules to statistical models (§2.4): abandon
the explicit encoding of expert understanding of problem structure into
programmatic rules, and instead use data---here, natural-language
reasoning chains---as the carrier of that judgment system.

\section{Experiments}\label{experiments}

\subsection{\texorpdfstring{Core Verification: Two Metrics --- \(\Delta(\Delta I)\) and \(\Delta\log P\)}{Core Verification: Two Metrics -- Delta(Delta I) and Delta log P}}\label{core-verification-two-metrics-deltadelta-i-and-deltalog-p}

\textbf{Benchmark composition.} The evaluation set consists of 80
independent benchmark items, split across four types: (1) 20 literary
judgment paired questions (each presents two candidate options for the
same narrative decision point, used to compute \(\Delta(\Delta I)\));
(2) 20 factual judgment paired questions (same format, serving as the
domain control group); (3) 20 literary continuation passages (using the
actual subsequent passage from a published work as ground truth \(x^*\),
for the \(\Delta\log P\) computation in §4.5); (4) 20 factual
continuation passages (for the cross-domain \(\Delta\log P\) comparison
in §4.5).

We apply a time-window control to all evaluation materials to reduce
contamination from pretraining memorization. The base model used in this
paper---Qwen2.5-7B-Base---has a pretraining data cutoff of June 2024
\citep{yang2024qwen25}. All ground-truth continuation passages were
individually verified by the authors to have been publicly published
after that cutoff date, to avoid explicit overlap with material the base
model may have seen.

\textbf{Three metrics:}

\begin{itemize}
\tightlist
\item
  \textbf{\(\Delta(\Delta I)\) (change in conditional mutual information
  --- primary metric):} For each of the 20 paired questions in the
  literary group and the 20 in the factual group, we compute
  \(\Delta I = I_{\text{good}}(X;Y) - I_{\text{bad}}(X;Y)\) separately
  for the base model (A) and the LoRA model (B), then report
  \(\Delta(\Delta I) = \Delta I_B - \Delta I_A\). Statistical tests:
  paired Wilcoxon signed-rank test (one-sided, \(H_1\): LoRA
  \textgreater{} base), paired \(t\)-test, and Cohen's \(d\).
\item
  \textbf{\(\Delta\log P\) per token (continuation --- auxiliary
  metric):} For the 20 literary continuation passages and 20 factual
  continuation passages, we report the per-token \(\log P\) difference
  between the base model and the LoRA model on the ground-truth passage
  \(x^*\) (see §4.5).
\item
  \textbf{Supervised eval\_loss (diagnostic only --- not used for model
  selection):} Token-level cross-entropy on the 11-sample validation
  set, reported as a diagnostic of the training process. In this
  experiment, this value rises monotonically from approximately 2.7 at
  epoch 1 to approximately 4.6 at epoch 12---a direction opposite to the
  target metric \(\Delta(\Delta I)\), which peaks at epoch 5. This
  confirms that token-CE is not a suitable model selection criterion in
  the small-validation-set setting (see §5.1 and §6.6).
\end{itemize}

\textbf{Model selection protocol.} \texttt{load\_best\_model\_at\_end}
is disabled throughout training. All 12 epoch checkpoints are saved. The
\(\Delta(\Delta I)\) evaluation is run at 7 sampling points (epochs
\(\{2, 3, 4, 5, 6, 8, 12\}\)), and epoch 5 is selected as the primary
result checkpoint based on the Wilcoxon \(p\)-value and Cohen's \(d\)
(rationale in §5.1).

\subsection{Comparison with Commercial State-of-the-Art}\label{comparison-with-commercial-state-of-the-art}

A natural extension of this work is to run parallel comparisons on the
same \(\Delta(\Delta I)\) evaluation materials against current
commercial state-of-the-art models (GPT-5.5-class, Claude Opus
4.7-class, Gemini 2.x-class). This requires normalized logprob scoring
across different tokenizers (using the rank probability of the correct
option within the candidate set as a unified metric---see discussion at
the end of §4.5) and stable commercial API access in long-context
settings. Due to submission timeline constraints, this comparison is
deferred to arXiv v2. The primary results in this paper focus on the
before-versus-after fine-tuning comparison on the same base model, to
isolate the marginal contribution of SNAKE-100 training data.

\subsection{Methodological Stance on Three-Dimension Ablation}\label{methodological-stance-on-three-dimension-ablation}

One of the central methodological claims in §1.3 is that all three types
of tacit knowledge must be present in the CoT corpus as a whole. A
natural test for this claim would be a post-hoc ablation: hold the
corpus size fixed, remove one knowledge type at a time, retrain, and
measure how much \(\Delta(\Delta I)\) drops.

This paper does not run that ablation, and does not commit to running it
in future work. Two reasons.

First, each SNAKE-100 CoT is a holistic judgment produced naturally in a
BC dialogue. At the moment the expert makes that judgment, all three
constraint types have already converged into a single intersection
judgment in the expert's cognition (§3.1). Reconstructing the CoT by
slicing it dimension by dimension is epistemologically the same as
``linearizing the intersection judgment into parallel marginal
projections''---exactly the protocol rejected in §1.1 and §3.5. Using a
protocol already shown to be mathematically flawed as the ablation tool
is methodologically inconsistent.

Second, the three-dimension co-presence claim does not depend on
ablation for support. It already has theoretical grounding from the
independent academic precedents in §1.4
\citep{abrams1953,buhler1934,eco1990,grabe1996}, methodological
grounding from the BC Protocol's probing design in §3.1, and indirect
empirical support from the domain-selective results in §5.1 (literary
group significant, factual group null).

This section therefore explicitly removes the ``three-dimension ablation
experiment'' from the paper's experimental agenda---it will not be run
here, and is not committed to for arXiv v2.

\subsection{Scale Effects}\label{scale-effects}

The scale effect analysis described in §3.4 (training subsets of 10, 30,
60, and 100 CoT samples) is intended to characterize the marginal return
curve of SNAKE samples. The full execution requires 4 independent
fine-tuning runs and 4 complete evaluations. Due to submission timeline
constraints, this is deferred to arXiv v2. Note that the epoch-wise
sweep reported in §5.1 (7 sampling points) already provides model
selection evidence along a different dimension at fixed data size: it
shows that \(\Delta(\Delta I)\) reaches a sweet spot around epoch 4--5
under \(n = 105\) training samples. This provides an initial bound on
``when saturation occurs at approximately 100 samples,'' to be answered
more systematically by the scale effect experiment in arXiv v2.

\subsection{External Hard Validation: Logprob Evaluation on Published Works}\label{external-hard-validation-logprob-evaluation-on-published-works}

This section provides a validation that is independent of the self-built
benchmark, to address the concern that ``the self-built benchmark shares
the same formal framework as the training objective, creating a circular
validation.''

\textbf{Experimental design.} We select a number of published
high-quality short stories and chapters (Chinese and English, covering
different authors and genres). For each, we take a key narrative passage
as ground truth \(x^*\) and the preceding context as \(y\). The base
model, the CQA fine-tuned model, and several current commercial
state-of-the-art models (GPT-5.5 / Claude Opus 4.7 / Gemini 2.x, etc.)
are each asked, given the same context \(y\), to directly read the
logprob \(\log P_\theta(x^* \mid y)\) on the actual published passage.
We report the logprob differences between models (\(\Delta\log P\),
normalized in bits per token).

\textbf{Why this is external.} The ground truth is not constructed by
this paper---it is the actual next passage written by the author. The
circular-validation concern (``the model performs well on its own
benchmark'') does not apply here: the evaluation target (logprob on a
real creative writing distribution) is neither defined by this paper nor
within the training data's coverage.

\textbf{Relationship to the duality argument (§6.5).} This experiment
falls directly on an appreciation-side observable, but by the
architectural duality argued in §6.5, that observable is equivalent to a
measure of generation-side calibration. If the 7B fine-tuned model's
\(\log P\) on published literary continuations is systematically higher
than the base model's---or even higher than GPT-5.5-class frontier
models---then that result is mathematically equivalent to: the
fine-tuned model assigns higher probability to producing outputs close
to the real distribution of high-quality literary passages.

\textbf{Sample size and controls.} 5--10 source texts; each with 1--3
key narrative passages as ground truth. Time-window control: results are
reported separately for works published before and after each tested
model's training cutoff, to exclude contamination from ``model already
memorized this work.''

\textbf{Main results.} Table 2 reports per-token \(\log P\) statistics
at the epoch 5 checkpoint across text domains.

\textbf{Table 2: Base vs.~LoRA Per-Token \(\log P\) (epoch 5)}

{\def\LTcaptype{none} 
\begin{longtable}[]{@{}
  >{\raggedright\arraybackslash}p{(\linewidth - 10\tabcolsep) * \real{0.1667}}
  >{\raggedright\arraybackslash}p{(\linewidth - 10\tabcolsep) * \real{0.1667}}
  >{\raggedright\arraybackslash}p{(\linewidth - 10\tabcolsep) * \real{0.1667}}
  >{\raggedright\arraybackslash}p{(\linewidth - 10\tabcolsep) * \real{0.1667}}
  >{\raggedright\arraybackslash}p{(\linewidth - 10\tabcolsep) * \real{0.1667}}
  >{\raggedright\arraybackslash}p{(\linewidth - 10\tabcolsep) * \real{0.1667}}@{}}
\toprule\noalign{}
\begin{minipage}[b]{\linewidth}\raggedright
Text Domain
\end{minipage} & \begin{minipage}[b]{\linewidth}\raggedright
\(n\)
\end{minipage} & \begin{minipage}[b]{\linewidth}\raggedright
Base Mean
\end{minipage} & \begin{minipage}[b]{\linewidth}\raggedright
LoRA Mean
\end{minipage} & \begin{minipage}[b]{\linewidth}\raggedright
\(\Delta\)
\end{minipage} & \begin{minipage}[b]{\linewidth}\raggedright
95\% CI
\end{minipage} \\
\midrule\noalign{}
\endhead
\bottomrule\noalign{}
\endlastfoot
Literary continuation & 20 & -2.890 & -3.703 & \textbf{-0.813} &
{[}-0.894, -0.731{]} \\
News (factual control) & 10 & -2.496 & -3.104 & -0.609 & {[}-0.762,
-0.456{]} \\
Popular science (factual control) & 10 & -2.300 & -3.032 & -0.732 &
{[}-0.943, -0.520{]} \\
Factual combined & 20 & -2.398 & -3.068 & -0.670 & {[}-0.791,
-0.549{]} \\
\end{longtable}
}

\textbf{Interpreting the results.} The LoRA model's per-token \(\log P\)
is lower than the base model's across all text domains (\(\Delta < 0\)),
with the literary domain showing a slightly larger decrease (-0.813)
than the factual domain (-0.670). This does not directly correspond to
the \(\Delta(\Delta I)\) results in §5.1---which show a significant
improvement in literary paired discrimination
(\(\Delta(\Delta I) = +9.58\)) and a null effect for the factual domain
(\(\Delta(\Delta I) = -3.04\)).

The two metrics measure different things. \(\Delta(\Delta I)\) measures
the model's discrimination sensitivity on \textbf{paired options} (the
degree to which the good option receives a higher \(\Delta I\) than the
bad option, given constraint \(Y\)). \(\log P\) measures the absolute
generation probability on \textbf{arbitrary text passages}. SFT
fine-tuning can improve paired discrimination sensitivity while
simultaneously narrowing the probability distribution overall---causing
absolute \(\log P\) to decrease across all domains.

Specifically: SFT constrains the model's probability mass toward the
judgment paradigm represented by the training data (SNAKE-100). This
concentration effect narrows the distribution uniformly at the absolute
logprob level, across all domains including the test-set literary
domain. It does not mean the model's discrimination ability on those
texts has decreased---that is what \(\Delta(\Delta I)\) captures, not
the absolute \(\log P\). The slightly larger literary decrease (-0.813
vs.~-0.670) is consistent with the training data being concentrated in
the literary domain.

This paper uses \(\Delta(\Delta I)\) (paired discrimination sensitivity)
as the primary calibration metric. \(\Delta\log P\) is presented as a
supplementary view. Its relationship to the duality argument in §6.5 is
discussed further at the end of §5.1. The comparison with commercial
SOTA models is planned for arXiv v2 (§4.2).

\section{Results}\label{results}

\subsection{Core Results}\label{core-results}

\textbf{Table 1: Epoch-wise \(\Delta(\Delta I)\) Summary} (literary
group \(n=20\); factual group \(n=20\))

{\def\LTcaptype{none} 
\begin{longtable}[]{@{}
  >{\raggedright\arraybackslash}p{(\linewidth - 12\tabcolsep) * \real{0.1429}}
  >{\raggedright\arraybackslash}p{(\linewidth - 12\tabcolsep) * \real{0.1429}}
  >{\raggedright\arraybackslash}p{(\linewidth - 12\tabcolsep) * \real{0.1429}}
  >{\raggedright\arraybackslash}p{(\linewidth - 12\tabcolsep) * \real{0.1429}}
  >{\raggedright\arraybackslash}p{(\linewidth - 12\tabcolsep) * \real{0.1429}}
  >{\raggedright\arraybackslash}p{(\linewidth - 12\tabcolsep) * \real{0.1429}}
  >{\raggedright\arraybackslash}p{(\linewidth - 12\tabcolsep) * \real{0.1429}}@{}}
\toprule\noalign{}
\begin{minipage}[b]{\linewidth}\raggedright
Epoch
\end{minipage} & \begin{minipage}[b]{\linewidth}\raggedright
Literary \(\Delta(\Delta I)\)
\end{minipage} & \begin{minipage}[b]{\linewidth}\raggedright
Literary Wilcoxon \(p\)
\end{minipage} & \begin{minipage}[b]{\linewidth}\raggedright
Literary \(d\)
\end{minipage} & \begin{minipage}[b]{\linewidth}\raggedright
Factual \(\Delta(\Delta I)\)
\end{minipage} & \begin{minipage}[b]{\linewidth}\raggedright
Factual \(p\)
\end{minipage} & \begin{minipage}[b]{\linewidth}\raggedright
Factual \(d\)
\end{minipage} \\
\midrule\noalign{}
\endhead
\bottomrule\noalign{}
\endlastfoot
2 & +1.54 & 0.273 & 0.22 & -1.37 & 0.727 & -0.14 \\
3 & +4.12 & 0.226 & 0.30 & -2.34 & 0.676 & -0.13 \\
4 & +7.87 & \textbf{0.049} & 0.47 & -1.64 & 0.551 & -0.07 \\
\textbf{5} & \textbf{+9.58} & \textbf{0.041} & \textbf{0.50} & -3.04 &
0.689 & -0.12 \\
6 & +8.41 & 0.095 & 0.38 & -0.97 & 0.580 & -0.04 \\
8 & +9.57 & 0.077 & 0.38 & -3.04 & 0.676 & -0.10 \\
12 & +11.71 & 0.062 & 0.39 & -1.15 & 0.536 & -0.03 \\
\end{longtable}
}

\begin{figure}[t]
  \centering
  \includegraphics[width=0.85\linewidth]{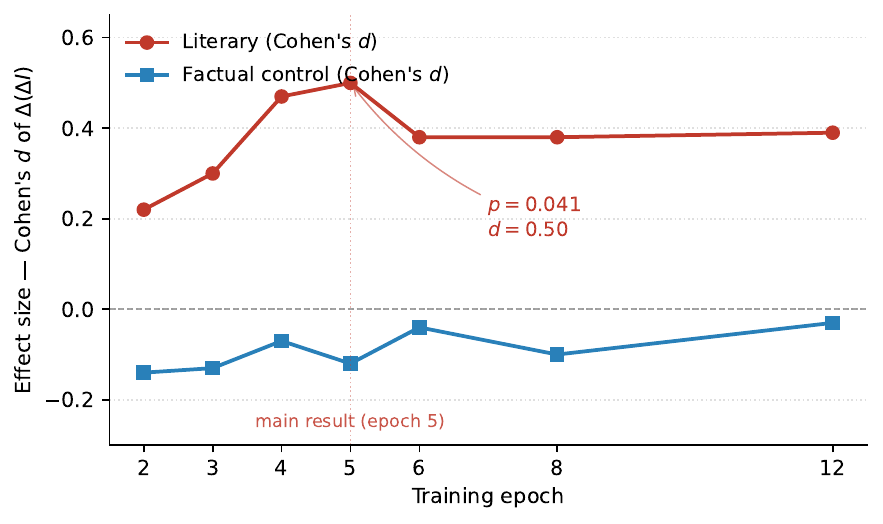}
  \caption{Domain selectivity. Cohen's $d$ of $\Delta(\Delta I)$ over training epochs:
    the literary group peaks at epoch~5 ($d=0.50$, $p=0.041$);
    the factual control remains near zero across all epochs.}
  \label{fig:domain-selectivity}
\end{figure}

At epoch 5: literary group \(\Delta(\Delta I) = +9.58\) (base mean
14.37, LoRA mean 23.94), Wilcoxon signed-rank one-sided test
\(p = 0.041\), Cohen's \(d = 0.50\). This is a medium effect size on
\(n = 20\) paired samples. At the same checkpoint, factual group
\(\Delta(\Delta I) = -3.04\), \(p = 0.689\), \(d = -0.12\)---no
significant effect, effect size near zero.

This contrast is the core empirical claim of the paper:
\textbf{SNAKE-100 fine-tuning systematically increases the LoRA model's
sensitivity to the full-dimensional constraint \(Y\) in literary
judgment (\(\Delta(\Delta I) > 0\), statistically significant), while
not inflating confidence on the factual judgment dimension} (factual
group \(d \approx 0\), \(p > 0.5\)).

The epoch-wise sweep (Table 1, Figure 1) shows that significance for the
literary group first appears at epoch 4 (\(p = 0.049\), \(d = 0.47\)),
peaks at epoch 5 (\(p = 0.041\), \(d = 0.50\)), and then---as training
continues---the mean effect size keeps rising (epoch 12:
\(\Delta(\Delta I) = +11.71\)) but between-sample variance also grows,
pulling the Wilcoxon \(p\)-value back to the \(0.06\)--\(0.10\) range.
This pattern matches the typical ``post-overfitting drop in sample
consistency'' behavior in small-data SFT. The factual group stays at
null effect across all epochs (all \(|d| < 0.15\)), ruling out the
confound of ``extended training causing global distributional drift.''

One important note: the eval\_loss on the 11-sample validation set rises
monotonically throughout training (from approximately 2.73 at epoch 1 to
approximately 4.62 at epoch 12). By conventional supervised learning
intuition this would be diagnosed as overfitting, warranting early
stopping. This paper nevertheless runs the full 12-epoch training and
reports epoch 5 (not epoch 1) as the primary result. The reason is an
explicit methodological judgment: \textbf{selecting a checkpoint by
token-CE on a small validation set is structurally misaligned with the
metrics this paper actually cares about (\(\Delta(\Delta I)\) and domain
selectivity)}. Token-CE measures how well the model can reproduce the
surface token sequence of expert CoT. \(\Delta(\Delta I)\) measures the
model's conditional distribution sensitivity to literary constraints.
These measure different things. Distortion on the former in a small
validation set should not drive model selection for the latter. This is
elaborated in §6.6.

\textbf{A precise statement on overfitting.} From the perspective of the
target metric \(\Delta(\Delta I)\), the 12-epoch training does not
exhibit typical overfitting. The literary group's mean effect size rises
continuously (from \(+9.58\) at epoch 5 to \(+11.71\) at epoch 12). The
factual group stays at null across all epochs. The only sense in which
``overfitting'' appears is that Wilcoxon significance drops from
\(p = 0.041\) at epoch 5 to \(p \approx 0.06\)--\(0.10\) afterward---due
to growing between-sample variance, not a reversal of the target metric.
This stands in sharp contrast to the picture suggested by monotonically
rising eval\_loss---and further confirms that token-CE is not a suitable
model selection criterion in this setting.

\subsection{Comparison Results}\label{comparison-results}

As stated in §4.2, comparison with commercial SOTA models is deferred to
arXiv v2.

\subsection{Status of Ablation Experiments}\label{status-of-ablation-experiments}

As stated in §4.3, this paper explicitly does not run post-hoc slice
ablations. Support for the three-dimension co-presence claim is provided
jointly by: the academic precedents in §1.4, the BC Protocol probing
design in §3.1, and the domain contrast results (literary vs.~factual)
in §5.1. The epoch-wise sweep (Table 1) and domain selectivity together
constitute the internal robustness case for the primary result.

\subsection{Scale Effect Analysis}\label{scale-effect-analysis}

As stated in §4.4, the 10/30/60/100 sample scale sweep is deferred to
arXiv v2. The epoch-wise \(\Delta(\Delta I)\) trajectory from §5.1
(Table 1) provides an initial constraint: \(\Delta(\Delta I)\) reaches a
sweet spot around epochs 4--5 and then shows growing between-sample
variance under \(n = 105\) training samples. This suggests the model's
discrimination sensitivity is already effectively calibrated at this
data size, with diminishing marginal returns from additional training. A
more systematic answer awaits the scale experiment in arXiv v2.

\section{Discussion}\label{discussion}

\subsection{The Conditional Entropy Perspective}\label{the-conditional-entropy-perspective}

The experimental results support the following reading: the difference
between an LLM with CQA fine-tuning and one without---on creative
judgment tasks---is concentrated in the conditional entropy \(H(X|Y)\).

\textbf{Without CQA fine-tuning.} Given the full-dimensional constraint
\(Y\), the model's conditional distribution \(P(x|y)\) still tends to
spread probability across multiple candidates, corresponding to a high
\(H(X|Y)\). This is not because the model lacks knowledge about the
relevant constraints. Pretraining corpora contain substantial text
covering psychology, causality, tone, and other related dimensions. The
issue is that ``having seen'' is not the same as ``being sensitive to'':
the model lacks the ability to sharpen its conditional distribution
based on the full-dimensional constraint in the context of specific
narrative decisions.

\textbf{With CQA fine-tuning.} Given the same constraint \(Y\), the
model can concentrate higher probability on the small set of candidates
that genuinely satisfy the full-dimensional constraint, corresponding to
a significantly reduced \(H(X|Y)\).

The unconditional entropy \(H(X)\)---the surprise of the choice when no
constraint is considered---remains largely unchanged before and after
CQA fine-tuning. What fine-tuning changes is \(H(X|Y)\), not \(H(X)\).

CQA can therefore be characterized as follows: it does not change the
model's overall prior distribution over the creative space. It changes
the sharpness of the set of plausible candidates the model can identify
after conditioning on the full-dimensional constraint.

This conditional distribution sharpening has empirical support in our
experiments. Table 1 and Figure 1 show that the literary group's
significant improvement in \(\Delta(\Delta I)\) (\(d = 0.50\),
\(p = 0.041\)) and the factual group's null effect on the same metric
(\(d = -0.12\), \(p = 0.689\)) hold simultaneously---constituting
empirical evidence that \(P(x|y)\) sharpening is
\textbf{domain-selective}.

\begin{figure}[t]
  \centering
  \includegraphics[width=0.85\linewidth]{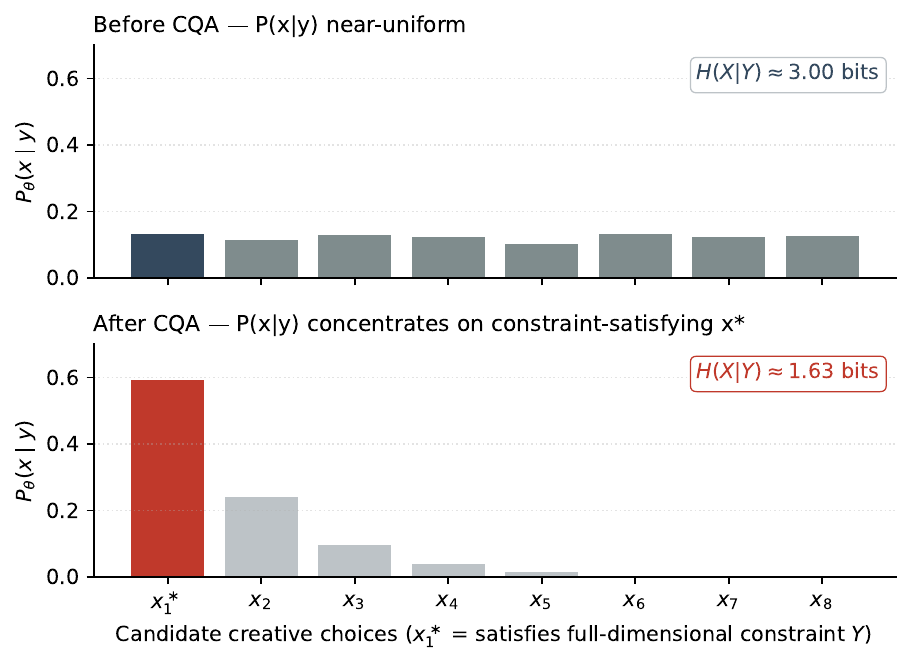}
  \caption{Conditional probability collapse after CQA fine-tuning.
    $P_\theta(x \mid y)$ concentrates on the constraint-satisfying
    candidate $x_1^*$, reducing $H(X \mid Y)$ while $H(X)$
    remains unchanged.}
  \label{fig:cond-entropy-collapse}
\end{figure}

\subsection{Theoretical Position of the Lossy Compression Perspective}\label{theoretical-position-of-the-lossy-compression-perspective}

The ``actively choose where information loss occurs'' strategy proposed
in §3.3 can be placed within the framework of Rate-Distortion Theory
from Shannon information theory for a more precise characterization.

The information carried by an expert's internal judgment system far
exceeds the capacity of a few hundred CoT samples. Any extraction
process toward natural language is, in principle, lossy. The
methodological distinction is not whether there is loss, but whether the
location of loss is actively chosen: passive loss means not knowing what
was dropped; active loss means using methodological judgment to choose
which information---the parts with relatively lower information
gain---to sacrifice.

The logprob-based prescreening strategy proposed in §3.3.1 is one
operational form of the latter: by directly measuring the pretraining
model's current calibration at each decision point, it concentrates the
CoT budget on the decision points where calibration is most needed.

\subsection{Position within the Research System}\label{position-within-the-research-system}

This paper sits alongside three companion papers in a coordinated
research line. The division of labor is as follows:

\textbf{``Calibrated Surprise'' \citep{calibratedsurprise2026}} provides
the formal mathematical definition of creative quality:
\(I(X;Y) = H(X) - H(X|Y)\). The \(H(X|Y)\) term is exactly what this
paper's experiments act on. The two lines of work correspond to each
other mathematically.

\textbf{BC Protocol paper \citep{bc2026}} provides the data production
pipeline. The natural-language CoT produced in BC dialogues is the
training material for CQA.

\textbf{Benchmark paper \citep{bench2026}} provides the independent
external diagnostic tool. The claim ``fine-tuning on small-scale expert
CoT can calibrate \(P(x|y)\) in a small base model'' requires external
comparative evidence from an independent benchmark. That verification is
handled by the Benchmark paper.

The three layers are not interchangeable: they operate on different
objects and serve different purposes.

\subsection{Applications: Paths from Research to Practice}\label{applications-paths-from-research-to-practice}

\textbf{From fiction to film, game NPCs, and brand storytelling---the
pipeline transfers.} The only domain-specific component in this paper's
engineering pipeline is the expert who provides the constraint
\(Y\)---the remaining components are domain-independent in principle.
From narrative fiction to film screenwriting, game NPC dialogue, brand
content creation, and music and visual arts: each domain only requires
replacing the expert with someone who carries the corresponding three
types of tacit knowledge. The CQA pipeline itself does not need to be
redesigned.

\textbf{One-time expert investment, industrial-scale
distribution---CQA's marginal economics.} A successfully calibrated
model has the property of ``injected once, distributed at near-zero
marginal cost.'' Top experts externalize their knowledge through a
single CoT production session; the model then distributes that judgment
capability to downstream applications at industrial scale. Notably, this
distribution does not require a separate training run for the generation
side: by the appreciation-generation duality argued in §6.5, calibrating
the appreciation side automatically transfers to the generation side.
One CoT investment, both sides benefit. No separate deployment track for
appreciation versus generation.

\textbf{Precision content recommendation---using \(I(X;Y)\) to measure
audience-work fit.} Within the same \(I(X;Y)\) formula, holding the work
fixed and varying the reader's \(Y\) makes it possible to quantify the
aesthetic match between different audience groups and a given work. When
anchored to \(P_{\text{expert}}(x)\), the measurement reflects the
work's intrinsic quality. When anchored to \(P_{\text{general}}(x)\), it
reflects the work's reception quality from that audience's perspective.
Both can be discussed within the same unified framework, providing a
theoretical foundation for precision content distribution, personalized
recommendation, and audience segmentation research.

\textbf{A formal starting point for next-generation content evaluation
pipelines.} This paper's framework identifies the structural limitations
of rubric decomposition at the mathematical level, and provides the
joint probability \(P(x|y_1, \ldots, y_n)\) path as a theoretical
reference point for alternative evaluation paradigms. This can serve as
a formal skeleton for industrial content moderation, creative quality
evaluation, and automatic content curation systems.

\textbf{An opening for collaboration with open-source model providers.}
This paper verifies \(P(x|y)\) calibration via logprobs---the most
complete observable of a model's internal state available through
standard APIs. For research institutions with full model weights, deeper
observation methods are available: activation probing can directly read
intermediate hidden-layer vector representations to observe how CQA
fine-tuning changes the model's internal representations; attention
pattern analysis can check whether the fine-tuned model actually
allocates attention to positions relevant to the full-dimensional
constraint. Such methods would push CQA verification from ``is the
conditional distribution calibrated'' to ``how is that calibration
implemented inside the model.'' This requires open-source models or
cooperation from model weight holders. The theoretical framework and
experimental design in this paper apply equally to such follow-up work.

\subsection{\texorpdfstring{Why Learning to ``Read'' a Work Means Learning to ``Write'' One: The Duality of a Single \(P_\theta\)}{Why Learning to Read Means Learning to Write: Duality of P-theta}}\label{why-learning-to-read-a-work-means-learning-to-write-one-the-duality-of-a-single-p_theta}

This section expands the remark at the end of §1.2---``\(H(X|Y)\)
characterizes both sides at once''---into a full argument. It responds
to a common methodological objection (``you didn't test generation; why
claim that creative judgment was calibrated?''), and explains why
$\sim$100 CoT examples are sufficient at the engineering level
without appealing to purely empirical observations like
LIMA~\citep{zhou2023lima}.

\textbf{1. Two forward passes of a single conditional distribution.} A
large language model is, mathematically, exactly one family of
conditional distributions \(P_\theta(x_t \mid x_{<t})\). All operations
on this model fall into exactly two categories:

\begin{itemize}
\tightlist
\item
  \textbf{Generation:} Sample the next token from
  \(P_\theta(x_t \mid x_{<t})\).
\item
  \textbf{Appreciation:} Read \(\log P_\theta\) over a given token
  sequence from \(P_\theta(x_t \mid x_{<t})\), obtaining the
  log-likelihood of that sequence under the current model.
\end{itemize}

These are not two independent models, two independent parameter sets, or
``two similar capabilities.'' They are \textbf{two forward passes of the
same \(P_\theta\)}---the same softmax output vector, used once as a
sampling distribution and once as a likelihood function. This structure
holds for all autoregressive LLMs, regardless of the specific
architecture (Transformer, SSM, etc.).

\textbf{2. Mutual information symmetry.} By Shannon information theory:

\[I(X;Y) \;=\; H(X) - H(X|Y) \;=\; H(Y) - H(Y|X)\]

In the context of this paper: the degree to which the creative choice
\(X\) is sharpened by constraint \(Y\)---and the degree to which
constraint \(Y\) is revealed by the creative stream \(X\)---are
\textbf{the same scalar}. There is no solution in which
appreciation-side precision is high while generation-side precision is
low, or vice versa. Such a solution would require two different
\(P_\theta\) objects to exist.

\textbf{3. The generative-discriminative duality.} A classical ML result
\citep{ng2001}: a perfect generative model can be translated into a
perfect discriminator at zero cost (just read the likelihood), and vice
versa. Asymmetries only appear under finite data and uneven task
difficulty. In this paper's setting, ``generation'' and
``discrimination'' share the same token-level probability space. This
duality collapses in principle into identity, not approximate equality.

\textbf{4. The human counterexample does not apply.} It is common to
observe ``strong appreciation, weak creation'' in humans---many heavy
readers have sharp aesthetic intuitions but cannot independently produce
work of comparable quality. This is not evidence that the two
capabilities are separable. It is evidence that \textbf{humans have two
different hardware systems}: appreciation runs on cheap, parallel,
System-1-style pattern recognition; creation requires expensive,
sequential processing that needs long working memory and motor planning.
These two systems can be trained independently, so one being strong and
the other weak is entirely possible in humans.

LLMs have no such architectural separation. Appreciation and generation
share the same weights, the same forward pass. \textbf{Therefore, in
LLMs, the human counterexample does not hold}---and this is precisely
the leverage that CQA exploits.

\textbf{5. Two direct corollaries for CQA.}

\emph{Corollary 1 (structural source of data efficiency):} Calibrating
the appreciation side is equivalent to calibrating the generation side.
CQA fine-tuning acts directly on the appreciation side's log-likelihood
at the loss level (targeting the expert CoT distribution), but its
effect transfers automatically to the generation side via architectural
duality. This means ``why $\sim$100 samples are enough'' has a
structural reason in the CQA framework---not a purely empirical
observation like LIMA~\citep{zhou2023lima}. The
traditional RLHF asymmetry (requiring large amounts of data to move the
generation side) is structurally bypassed in a single-\(P_\theta\)
architecture. This argument does not replace the empirical evidence from
the scale effect experiments (§4.4). It provides an independently
checkable theoretical prediction for that empirical phenomenon.

\emph{Corollary 2 (experimental design tradeoff):} This paper uses
logprobs as the only externally observable quantity, and does not run
generation-side experiments (``let the model write a complete work for
human evaluation'') in the main paper. This tradeoff might previously
have looked like avoiding generation-side empirical evidence. The
duality argument reverses that reading: in a single-\(P_\theta\)
architecture, systematic changes in appreciation-side logprobs are
equivalent to systematic changes in generation-side capability. The two
do not need to be measured separately. One caveat: this equivalence
holds at the internal distribution level. The generation side can still
be affected at the decoding strategy level (temperature, top-p, beam
search, etc.). Those ``decoding-level'' differences are outside the
scope of this paper's calibration target.

\textbf{6. Scope conditions and future work.} This duality argument
depends on two premises: (1) the model is a single autoregressive
conditional distribution, with no separate discriminator head or reward
model; (2) generation and appreciation are performed under the same
context format (otherwise the two forward passes operate on different
conditional distributions, and the symmetry no longer applies directly).
Extensions to multi-model routing architectures, Mixture-of-Experts with
asymmetric expert usage, and external reward model settings are left for
future work. All experiments in this paper are conducted under settings
that satisfy both premises (§3.4).

This argument can also be probed empirically in the external hard
validation of §4.5 (directly measuring how the fine-tuned model's
logprob on actual published continuations differs from the base
model's)---that experiment falls on an appreciation-side observable
that, by duality, translates into equivalent evidence about
generation-side calibration.

\subsection{Limitations}\label{limitations}

\textbf{Sample size.} This paper uses approximately 100 CoT samples as
the training data. Their adequacy should be addressed by the scale
effect analysis designed in §4.4. Additionally, as noted in §3.1, the
precision of B's stopping point on each sample has a hidden influence on
the effective information density of the dataset. Systematically
quantifying that influence is beyond this paper's scope.

\textbf{Domain generalization.} Experiments are conducted only within
the narrative fiction domain. Whether this paper's pipeline is effective
in other creative domains requires independent validation by re-running
the same pipeline in those domains.

\textbf{Model generalization.} This paper fine-tunes a 7B-parameter base
model. Whether the same pipeline works equally well on larger or smaller
base models requires further experiments.

\textbf{Granularity coverage.} This paper's experiments focus on
narrative decision-level choices (plot direction, character responses,
scene design, and other macro-level creative decisions). Shannon
information theory's chain rule imposes no mathematical constraint on
decision granularity. Whether micro-level craft decisions (sentence
rhythm, rhetorical strategy, humor timing, etc.) can be covered by the
same framework is left for future work.

\textbf{Validation set size and model selection metric mismatch.} This
paper uses an 11-sample validation set (held out from the 105 training
samples at a 0.1 ratio). At this size, token-level cross-entropy
eval\_loss is structurally misaligned with the target metric
\(\Delta(\Delta I)\)---the former rises monotonically throughout
training (approximately 2.73 → 4.62), while the latter peaks at epoch 5.
This paper's engineering solution is to disable
\texttt{load\_best\_model\_at\_end} and perform post-hoc model selection
based on \(\Delta(\Delta I)\). The more fundamental fix---using a
task-relevant metric for in-loop validation during training---is left
for future work.

\textbf{No blind evaluation of open-ended generation.} The primary
results rely on objective metrics that can be observed directly from
logprobs (\(\Delta(\Delta I)\) and \(\Delta\log P\)). The paper does not
include human or LLM-as-Judge blind evaluation on open-ended generation
tasks (such as writing literary commentary or continuations). This is
because in the small-model + LoRA + $\sim$100-sample setting,
the stylistic features of the training data impose a strong template
effect on the generation side, making it difficult to cleanly separate
``cognitive calibration'' from ``stylistic replication'' in generation
outputs. This decoupling problem is treated as a separate follow-up
task. Its effect on the main conclusion has a structural backstop from
the duality argument in §6.5: systematic appreciation-side logprob
changes are equivalent to systematic generation-side calibration
changes.

\section{Conclusion}\label{conclusion}

This paper provides empirical engineering support for the mathematical
claim that ``creative quality improves as \(H(X|Y)\) is reduced,''
proposed in \emph{Calibrated Surprise}. To make this support as
generalizable as possible, we deliberately choose the strictest
engineering conditions: approximately 100 expert chain-of-thought
samples produced by the BC Protocol, Qwen2.5-7B-Base as the base model,
and LoRA-based supervised fine-tuning.

The model's \(\Delta(\Delta I)\) on an independent benchmark shows a
significant reduction in \(H(X|Y)\). At the epoch 5 checkpoint: literary
group \(\Delta(\Delta I) = +9.58\) (\(p = 0.041\), Cohen's \(d = 0.50\),
\(n = 20\) paired); at the same checkpoint, factual judgment
\(\Delta(\Delta I) = -3.04\) (\(p = 0.689\), \(d = -0.12\))---no
significant effect. This contrast is direct empirical evidence that
``\(P(x|y)\) calibration is domain-selective,'' independent of human
subjective judgment and independent of the model's own verbal scoring.

The experiments also show that all three constraint types---author
intent, audience expectation, reality logic---must be present across the
CoT corpus as a whole. This claim maps onto a diagnosable data bias: in
the open alignment datasets surveyed, coverage is systematically skewed
toward one constraint type; the other two are systematically
underrepresented.

We use the term Creative Quality Alignment (CQA) to refer to this class
of engineering methods---those that extend from \emph{Calibrated
Surprise} toward aligning LLM creative judgment. This paper describes
one specific path within that class. Other engineering implementations
from the same theoretical base also belong to CQA.

To summarize: this paper establishes two independent lines of evidence
simultaneously. Domain selectivity (literary group significant, factual
group null) moves the calibratability of \(P(x|y)\) from theoretical
assumption to empirical fact. The skewed coverage of two of the three
constraint types in public alignment data is identified, for the first
time, in a diagnosable form. The foundational alignment work was
provided separately by \emph{Calibrated Surprise} and the BC Protocol
paper. This paper brings those two lines together, for the first time,
into a concrete path that can be verified by an independent benchmark.

\bibliography{references}

\begin{thebibliography}{21}
\providecommand{\natexlab}[1]{#1}
\providecommand{\url}[1]{\texttt{#1}}
\expandafter\ifx\csname urlstyle\endcsname\relax
  \providecommand{\doi}[1]{doi: #1}\else
  \providecommand{\doi}{doi: \begingroup \urlstyle{rm}\Url}\fi

\bibitem[Abrams(1953)]{abrams1953}
M.~H. Abrams.
\newblock \emph{The Mirror and the Lamp: Romantic Theory and the Critical
  Tradition}.
\newblock Oxford University Press, 1953.

\bibitem[B{\"u}hler(1934)]{buhler1934}
Karl B{\"u}hler.
\newblock \emph{Sprachtheorie: Die Darstellungsfunktion der Sprache}.
\newblock Fischer, 1934.
\newblock Organon Model.

\bibitem[Eco(1990)]{eco1990}
Umberto Eco.
\newblock \emph{The Limits of Interpretation}.
\newblock Indiana University Press, 1990.

\bibitem[Grabe and Kaplan(1996)]{grabe1996}
William Grabe and Robert~B. Kaplan.
\newblock \emph{Theory and Practice of Writing: An Applied Linguistic
  Perspective}.
\newblock Longman, 1996.

\bibitem[Howcroft et~al.(2020)Howcroft, Belz, Clinciu, et~al.]{howcroft2020}
David~M. Howcroft, Anja Belz, Miruna-Adriana Clinciu, et~al.
\newblock Twenty years of confusion in human evaluation: {NLG} needs evaluation
  sheets and standardised definitions.
\newblock In \emph{Proceedings of INLG 2020}, 2020.

\bibitem[Hu et~al.(2022)Hu, Shen, Wallis, et~al.]{hu2022lora}
Edward~J. Hu, Yelong Shen, Phillip Wallis, et~al.
\newblock {LoRA}: Low-rank adaptation of large language models.
\newblock In \emph{Proceedings of ICLR 2022}, 2022.

\bibitem[Hu et~al.(2024)Hu, Gao, Hu, Zhang, Chen, Xu, and Wan]{hu2024llm}
Xinyu Hu, Mingqi Gao, Sen Hu, Yang Zhang, Yicheng Chen, Teng Xu, and Xiaojun
  Wan.
\newblock Are {LLM}-based evaluators confusing {NLG} quality criteria?
\newblock In \emph{Proceedings of the 62nd Annual Meeting of the Association
  for Computational Linguistics (Volume 1: Long Papers)}, pages 9530--9555,
  2024.
\newblock ACL Anthology: 2024.acl-long.516.

\bibitem[Jain et~al.(2023)Jain, Chiang, Wen, et~al.]{jain2023neftune}
Neel Jain, Ping-yeh Chiang, Yuxin Wen, et~al.
\newblock {NEFTune}: Noisy embeddings improve instruction finetuning.
\newblock In \emph{Proceedings of ICLR 2024}, 2023.

\bibitem[Kalajdzievski(2023)]{kalajdzievski2023}
Damjan Kalajdzievski.
\newblock A rank stabilization scaling factor for fine-tuning with {LoRA},
  2023.
\newblock arXiv:2312.03732.

\bibitem[Kaufman et~al.(2008)Kaufman, Baer, Cole, and Sexton]{kaufman2008}
James~C. Kaufman, John Baer, Jason~C. Cole, and Janel~D. Sexton.
\newblock A comparison of expert and nonexpert raters using the consensual
  assessment technique.
\newblock \emph{Creativity Research Journal}, 20\penalty0 (2):\penalty0
  171--178, 2008.

\bibitem[Messick(1994)]{messick1994}
Samuel Messick.
\newblock The interplay of evidence and consequences in the validation of
  performance assessments.
\newblock \emph{Educational Researcher}, 23\penalty0 (2):\penalty0 13--23,
  1994.

\bibitem[Nathan and Petrosino(2003)]{nathan2003}
Mitchell~J. Nathan and Anthony Petrosino.
\newblock Expert blind spot among preservice teachers.
\newblock \emph{American Educational Research Journal}, 40\penalty0
  (4):\penalty0 905--928, 2003.

\bibitem[Ng and Jordan(2001)]{ng2001}
Andrew~Y. Ng and Michael~I. Jordan.
\newblock On discriminative vs.\ generative classifiers: A comparison of
  logistic regression and naive {Bayes}.
\newblock In \emph{Advances in Neural Information Processing Systems},
  volume~14, 2001.

\bibitem[Polanyi(1966)]{polanyi1966}
Michael Polanyi.
\newblock \emph{The Tacit Dimension}.
\newblock Doubleday, 1966.

\bibitem[Wang et~al.(2023)Wang, Kordi, Mishra, et~al.]{wang2023selfinstruct}
Yizhong Wang, Yeganeh Kordi, Swaroop Mishra, et~al.
\newblock Self-instruct: Aligning language models with self-generated
  instructions.
\newblock In \emph{Proceedings of ACL 2023}, 2023.

\bibitem[Yang et~al.(2024)Yang, Yang, Hui, et~al.]{yang2024qwen25}
An~Yang, Baosong Yang, Binyuan Hui, et~al.
\newblock Qwen2.5 technical report, 2024.
\newblock arXiv:2412.15115.

\bibitem[Zheng et~al.(2024)Zheng, Zhang, Zhang, et~al.]{zheng2024llamafactory}
Yaowei Zheng, Richong Zhang, Junhao Zhang, et~al.
\newblock {LLaMA-Factory}: Unified efficient fine-tuning of 100+ language
  models.
\newblock In \emph{Proceedings of ACL 2024 System Demonstrations}, 2024.

\bibitem[Zhou et~al.(2023)Zhou, Liu, Xu, et~al.]{zhou2023lima}
Chunting Zhou, Pengfei Liu, Puxin Xu, et~al.
\newblock {LIMA}: Less is more for alignment.
\newblock In \emph{Proceedings of NeurIPS 2023}, 2023.

\bibitem[Zou and Xu(2026{\natexlab{a}})]{bc2026}
Bo~Zou and Chao Xu.
\newblock {BC} protocol: Structured dual-expert dialogue for eliciting
  high-quality chain-of-thought post-training data, 2026{\natexlab{a}}.
\newblock Preprint. arXiv ID to be assigned.

\bibitem[Zou and Xu(2026{\natexlab{b}})]{bench2026}
Bo~Zou and Chao Xu.
\newblock Benchmark paper: Independent external diagnostic tool for {CQA}
  calibration verification, 2026{\natexlab{b}}.
\newblock Preprint. arXiv ID to be assigned.

\bibitem[Zou and Xu(2026{\natexlab{c}})]{calibratedsurprise2026}
Bo~Zou and Chao Xu.
\newblock Calibrated surprise: An information-theoretic account of creative
  quality.
\newblock \url{https://arxiv.org/abs/2604.26269}, 2026{\natexlab{c}}.
\newblock arXiv:2604.26269.

\end{thebibliography}

\end{document}